%% file: bare_conf.tex
\documentclass[conference]{IEEEtran}
\ifCLASSINFOpdf
   \usepackage[pdftex]{graphicx}
\else
\fi
\usepackage{amsmath, amssymb}
\DeclareMathOperator{\EX}{\mathbb{E}}

\usepackage{hyperref}

\usepackage{cleveref}                   

\usepackage{multirow}

\newcommand\ChangeRT[1]{\noalign{\hrule height #1}}

\usepackage{caption}

\begin{document}
%
\title{Autoencoders as Tools for Program Synthesis}

\author{\IEEEauthorblockN{Sander D.M. de Bruin}
\IEEEauthorblockA{s.d.bruin@hotmail.com\\
Eindhoven University of Technology\\
Eindhoven, The Netherlands\\}
\and
\IEEEauthorblockN{Vadim Liventsev}
\IEEEauthorblockA{v.liventsev@tue.nl\\
Eindhoven University of Technology\\
Eindhoven, The Netherlands}
\and
\IEEEauthorblockN{Milan Petkovic}
\IEEEauthorblockA{m.petkovic@tue.nl\\
Eindhoven University of Technology\\
Eindhoven, The Netherlands}}


%


\maketitle

\begin{abstract}
Recently there have been many advances in research on language modeling of source code. Applications range from code suggestion and completion to code summarization. However, complete program synthesis of industry-grade programming languages remains an open problem. In this work, we introduce and experimentally validate a variational autoencoder model for program synthesis of industry-grade programming languages. 
This model makes use of the inherent tree structure of code and can be used in conjunction with gradient free optimization techniques like evolutionary methods to generate programs that maximize a given fitness function, for instance, passing a set of test cases.
A demonstration is avaliable at \url{https://tree2tree.app}
\end{abstract}


%
\IEEEpeerreviewmaketitle

\section{Introduction}
\input{introduction}

\section{Background}
\input{background}

\section{Related work}
\input{related_work}



\section{Proposal: Tree-to-Tree}
\input{Tree2Tree}

\section{Experiments}
\input{Experiments}

\section{Conclusion}
\input{conclusion}





%


\bibliographystyle{abbrv}
\bibliography{references}

\end{document}

%% file: introduction.tex
There is a large number of applications for machine learning on programming languages, including but not limited to code suggestion and completion \cite{Luan_2019,li2018code}, program translation \cite{chen2018treetotree}, program repair and bug detection \cite{le2018maximal,hajipour2019samplefix}, code optimization \cite{bunel2017learning} and program synthesis \cite{shin2019program}. 
While current research on program synthesis has achieved impressive results \cite{gulwani2017program}, it often focuses on domain-specific languages \cite{liskowski2020program} or only considering a sub-domain of a programming language used in the software development industry. 

However, the main advantages of program synthesis stem from the fact that it allows for exchange of knowledge between human experts and data-driven machine learning models. 
Such an exchange occurs when:
\begin{enumerate}
    \item A program synthesis system uses software developed by people as a starting point in the optimization process, suggesting potential improvements. This task is known as genetic improvement of software \cite{genimp1,genimp2}
    \item Human developers can read the generated program and understand its approach to solving the problem.
\end{enumerate}

These advantages can only be achieved if the program synthesis model is using a programming language that human developers are widely familiar with.
According to TIOBE Index \cite{tiobe2017tiobe} these are language like Python, Java, C++ and JavaScript.
These languages are extremely sparse optimization spaces, meaning that a random string is highly unlikely to be a valid Python, Java or C++ program which makes it very hard to apply evolutionary techniques such as genetic programming \cite{turing1950computing,koza1992genetic}.

Genetic programming techniques maximize some metric of program quality (fitness function), such as how many unit tests the program passes, via the following loop:
\begin{enumerate}
    \item Extend the population of programs by applying random perturbation to the existing population
    \item Shrink the population of programs by discarding ones with the lowest fitness
\end{enumerate}

This approach is predicated on the assumption that some programs in the population have higher fitness than others, however in a sparse programming language this assumption is unlikely to hold: if most (or, worse, all) programs in the population don't compile, they all have an equal fitness and discarding lowest-fit programs does not lead to improvement.

To mitigate this issue, genetic programming has to be applied not in the space of code string, but in a subspace that only or mostly contains valid programs.
In this paper we introduce a method of inducing such a subspace by:
\begin{enumerate}
    \item Generating Abstract Syntax Trees as opposed to code strings. Code strings can contain errors that prevent them from being parsed by the compiler and represented as an AST are thus excluded from the optimization space \cite{ast}.
    \item Training an autoencoder model \cite{li2015hierarchical} on a corpus of ASTs. The latent space of such an autoencoder does not contain types of programs that never occur in the real world corpus, further restricting the optimization space.
\end{enumerate}

The result is a latent space where any random vector is mapped to a realistic program. 
Evolutionary search in this latent space is thus a much more tractable problem than genetic programming in the space of code strings.

%% file: background.tex
\label{sec:VAE}
The Variational Autoencoder (VAE) \cite{kingma2013auto}, is a machine learning architecture that provides a way to train a generative model with some prior distribution $p(z)$ and a neural network that can generate a sample given $z$, which we can denote as $p(x|z)$. Using simple probability rules, we can specify our maximization target in terms of these two steps:

\begin{equation}\label{eq:vae_first_eq}
    p(x) = \int p(x,z)dz = \int p(x|z)p(z)dz
\end{equation}

However, optimizing this target is not always tractable due to the integral not having an analytical solution \cite{kingma2019introduction}. Instead, let us introduce a inference model $p(z|x)$ which maps $x$ to $z$. We want to sample latent values from $x$ and introduce a learnable auxiliary distribution $q(z|x)$ that approximates $p(z|x)$, to overcome the intractability. We would then like to optimize the auxiliary distribution such that:

\begin{equation}
    q(z|x) \approx p(z|x)
\end{equation}

Using this new auxiliary distribution, we can rewrite the intractable target function:

\begin{equation}
    \log p(x) \geq \EX_{q(z|x)} \left[\log p(x|z)\right] - D_{KL}\big(q(z|x)||p(z)\big)\label{eq:ELBO}.
\end{equation}

We are now expressing a lower bound (ELBO) on $\log p(x)$ by rewriting our target function. \Cref{eq:ELBO} contains the Kullback-Leibner divergence, denoted by $D_{KL}$, which is a measure of how similar two distributions are \cite{kullback1951information}. Typically, the prior $p(z)$ is chosen to be a standard multivariate Gaussian with zero mean and diagonal unitary co-variance \cite{takahashi2019variational}. Let us consider that we are representing $q(z|x)$ with a neural network that computes the mean and diagonal unitary co-variance for a Gaussian distribution. The KL divergence term can be computed analytically in closed form:

\begin{equation}
    D_{KL} = \frac{1}{2} \sum_i \left(\sigma_i^2(x) + \mu_i^2(x) - \log \sigma_i^2(x) - 1\right)\label{eq:KL_div}
\end{equation}

where $n$ is the dimensionality of the multivariate Gaussians $p(z)$ and $q(z|x)$. Furthermore we can use the reparameterization trick \cite{kingma2013auto} to overcome the intractability of the log-likelihood term:

\begin{equation}
    \EX_{q(z|x)} \left[\log p(x|z)\right] = \EX_{p(\epsilon)} \left[\log p(x|z=\mu + \sigma \cdot \epsilon)\right]
\end{equation}

where $\mu$ and $\sigma$ are outputs of the neural network represented by $q(z|x)$, and $p(\epsilon)$ is a standard multivariate Gaussian distribution.

With these techniques, the optimization target becomes tractable. This allows us to compute gradients, if we sample $\epsilon$ and treat it as a constant, and use a stochastic gradient descent optimizer to update the model's parameters and train a neural network optimizing the target function.

One way to interpret the VAE is that it is a modification to the standard autoencoder where would like to learn a mapping $x \rightarrow z$ from the input data to a compressed representation, and the inverse of this mapping $z \rightarrow x$ to reconstruct the input data from the compressed representation. In the VAE, we replace the deterministic mapping $x \rightarrow z$ with a probabilistic network $q(z|x)$ and employ regularisation to ensure that this distribution is close to a predefined prior $p(z)$. This regularisation introduces structure into the latent space because VAEs encode the input as a distribution over the latent space, which allows for the generation of new samples.

%% file: related_work.tex


Liskowski \textit{et al.} \cite{liskowski2020program} learn program embedding by representing programs as abstract syntax trees, which are mapped to a low-dimensional continuous latent vector, and then back to abstract syntax trees. This embedding is learned via a variational RNN (LSTM \& GRU) autoencoder model. The embedding is then used to traverse the space of programs by a continuous optimization algorithm to find an optimal program according to the predefined goal. The result of the paper shows how program synthesis can be effectively rephrased as a continuous optimization problem using embeddings.

The goal of \cite{liskowski2020program} is essentially equivalent to our goal. However, the program domains that are considered are much simpler than what we would like to consider. The program synthesis tasks considered use very small token vocabulary sizes (i.e., the largest vocabulary size is 17), by working with, for example, the Boolean domain. In contrast, we would like to consider more popular, industry-grade programming languages (e.g., C++, Python). These programming languages are more complex and have much larger vocabulary sizes. Due to the `simple' program synthesis tasks, the proposed RNN autoencoder consists of only one or two layers in the encoder and decoder. We want to expand and improve on the structure of the models considered to accommodate much larger vocabulary sizes.

%% file: Tree2Tree.tex
 This section proposes an autoencoder model tailored for program synthesis. We will first discuss the design choices for the model. Afterward, we will elaborate on the architectures of the encoder and decoder models. Then, we will explain how the model will be optimized. Lastly, we will discuss how the data is pre-processed to fit the description of the model.

\subsection{Autoencoder type}
For our model, we have chosen to use a variational autoencoder (see \cref{sec:VAE}) since, unlike its non-stochastic counterpart,it is less dependent on choosing the right size of the latent vector, since the KL component encourages the model to use as small of a subspace of the latent space as it can.
Our experiments do, however, indicate that the choice of latent dimension is still important.

\subsection{Code representation}
Unlike natural languages, programming languages are easier to represent structurally due to the nature of their syntax which improves machine learning performance \cite{alon2019structural}\cite{chen2018tree}\cite{kusner2017grammar} compared to a more traditional sequence of tokens representation.
We are proposing a model that takes as input a tree and outputs a tree and will refer to our proposed autoencoder as the Tree-to-Tree (Tree2Tree) model.

\subsection{Encoder}
The encoder network aims to capture the most relevant information in a program and map it to a smaller representation.

\paragraph{Embedding} The first step when dealing with language is to convert tokens into dense representations, commonly referred to as embeddings. The first layer of the encoder network consists of an embedding layer, which can be either initialized randomly or initialized with pre-trained parameters and then fine-tuned further.

\paragraph{Tree-LSTM} As mentioned, we would like to take advantage of the structural information present in source code. There are two methods in which we could incorporate tree-structured data in the encoder network. The first method would be to flatten tree-structured data into a sequence and have the encoder operate on the flattened tree representation. There are multiple well-researched models capable of processing sequences, such as RNNs, Transformers, or CNNs. One of these established models could be used to process such a flattened tree representation. However, we want to operate directly on the tree representation to take advantage of the structural information. To allow for this, we require a model that can process tree-structured data. Tai \textit{et al.}\cite{tai2015improved} propose such a network by altering the standard LSTM architecture to enable capturing structural information. Instead of computing the hidden state from the input at the current time and the hidden state from the previous step, like in the standard LSTM model, the proposed model computes its hidden state from input at the current step and the hidden states from an arbitrary number of children.

We employ the Child-Sum Tree-LSTM \cite{tai2015improved} which is defined as follows. Given some tree, we can denote the set of children of a node $y$ as $C(y)$ and the vector representation of the node as $\mathbf{x}^y$. The transition equations between the different Tree-LSTM are the following:

\begin{align}\label{eq:tree_lstm_encocer}
    \mathbf{h}^y_* &= \sum_{z \in C(y)} \mathbf{h}^z \\
    \mathbf{i}^y &= \sigma(\mathbf{W}_{i} \cdot \mathbf{x}^y + \mathbf{U}_{i} \cdot \mathbf{h}^y_* + \mathbf{b}_{i})  \\ 
    \mathbf{f}^{yz} &= \sigma(\mathbf{W}_{f} \cdot \mathbf{x}^y + \mathbf{U}_{f} \cdot \mathbf{h}^z + \mathbf{b}_{f}) \\\label{eq:child_sum_4}
    \mathbf{o}^y &= \sigma(\mathbf{W}_{o} \cdot \mathbf{x}^y + \mathbf{U}_{o} \cdot \mathbf{h}^y_* + \mathbf{b}_{o}) \\
    \mathbf{u}^y &= tanh(\mathbf{W}_{u} \cdot \mathbf{x}^y + \mathbf{U}_{u} \cdot \mathbf{h}^y_* + \mathbf{b}_{u}) \\
    \mathbf{c}^y &= \mathbf{i}^y \odot \mathbf{u}^y + \sum_{z \in C(y)} \mathbf{f}^{yz} \odot \mathbf{c}^z \\
    \mathbf{h}^y &= \mathbf{o}^y \odot tanh(\mathbf{c}^y)
\end{align}

In \cref{eq:child_sum_4}, $z \in C(y)$ and $\odot$ denotes the element-wise product (Hadamard product). Furthermore, $W$, $U$ and $b$ denote trainable parameters of the model. Here we see that for calculating node $y$, we need to have computed the hidden states of all of the children $C(y)$. Hence, the computation order of the Tree-LSTM, given some tree, is bottom-up. Just like with the standard LSTM model, we can stack the Tree-LSTM to create a multilayer Tree-LSTM. In such a multilayer architecture, the hidden state of a Tree-LSTM unit in layer $l$ is then used as input to the Tree-LSTM unit in layer $l + 1$ in the same time step, the same as with the standard LSTM \cite{graves2013hybrid}. The idea is to let higher layers capture longer-term dependencies of the input. In the case of Tree-LSTMs, this translates to capturing longer-term dependencies along the paths of a tree.

\begin{figure}[ht!]
    \centering
    \includegraphics[width=\linewidth]{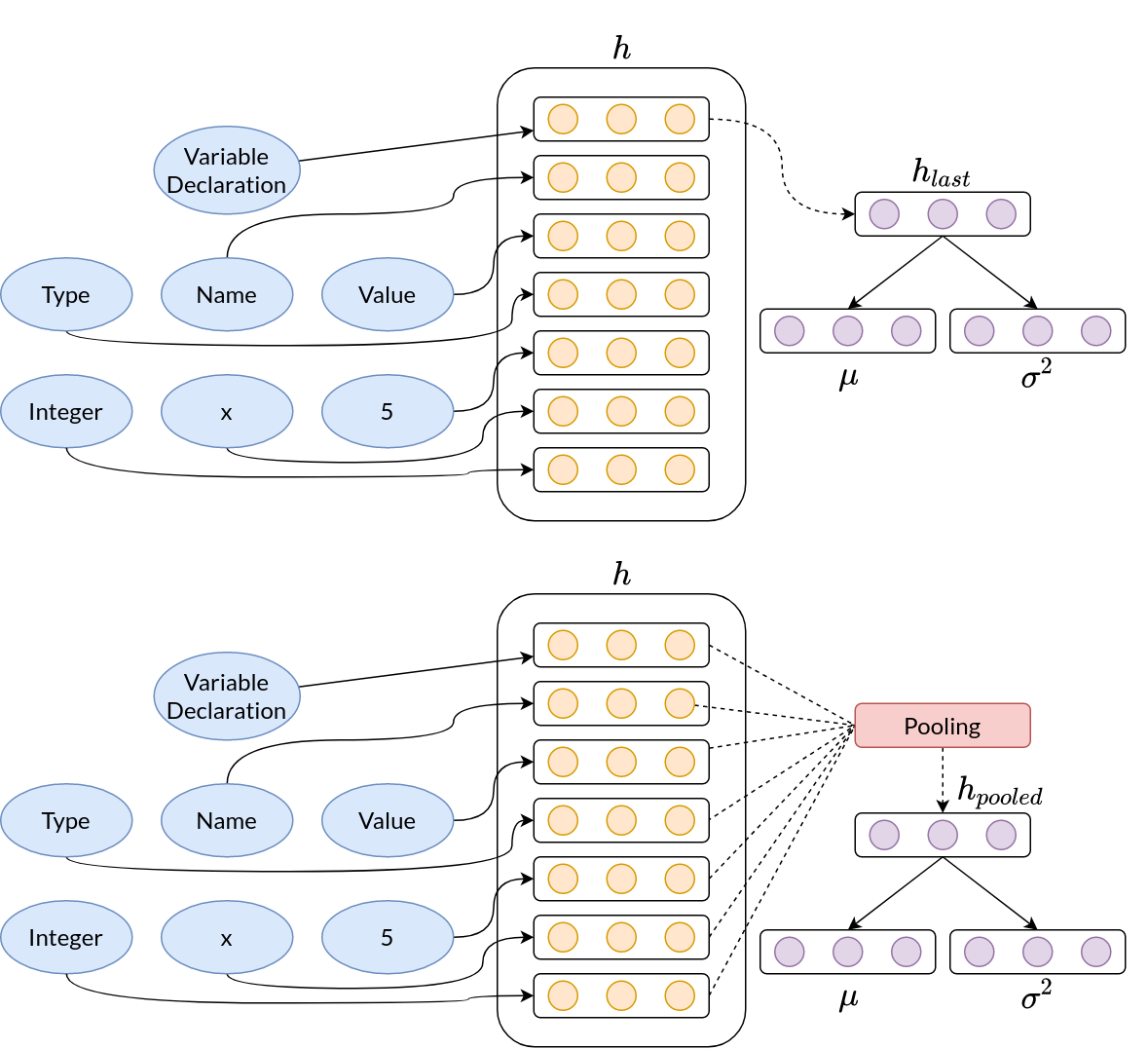}
    \caption[RNN pooling]{\textbf{Top}: Typical architecture of encoder model of VAE in which only the last hidden state from the RNN is used to compute the mean $u$ and variance $\sigma^2$. \textbf{Bottom}: A pooling method to aggregate the hidden states from the RNN to compute the mean $u$ and variance $\sigma^2$.}
    \label{fig:pooling}
\end{figure}

\paragraph{Neural attention} As the aim of the encoder network is to extract the most relevant information to a compressed state; we add an attention layer that comes after the last Tree-LSTM layer. This layer is fed with all the hidden states of the last Tree-LSTM layer. We add this layer because some nodes in the tree might contain more information to contribute to the latent code. The node importance calculation is based on \cite{winata2018attention}, and updates the hidden states as follows:

\begin{align}
    \mathbf{h}_{at} = \mathbf{h} \cdot tanh(\mathbf{W} \cdot \mathbf{h} + \mathbf{b})
\end{align}

Here, $h$ denotes the hidden states of the last Tree-LSTM layer. This additional layer allows the network to learn how to alter the Tree-LSTM layer(s) output to focus more on the nodes that contain the most information.

\paragraph{Pooling} We require our model to output a fixed size vector representing our latent code, however, the size of $h_{at}$ is variable as it is dependent on the size of the input. One approach to obtain a fixed size vector is to take the last hidden state after processing the entire tree. This method is often applied in autoregressive autoencoders \cite{fabius2015variational}. The intuition is that the last hidden state captures the entire input and summarizes this input into a single state. However, this method may suffer from long-term memory loss \cite{kao2020comparison}. Instead, pooling is a method to obtain a fixed size vector from the hidden states by aggregating features across different time steps of the hidden states. \cref{fig:pooling} depicts the difference between taking the last hidden state and pooling. Our model uses max-pooling to obtain a fixed size vector from our hidden states.

\paragraph{Sampling latent code} The pooled vector is then used to compute the mean and variance of the approximate posterior to sample a latent code $z$ with the help of the reparametrization trick \cite{kingma2013auto}. The mean and variance are computed using linear layers that learn a set of weights and biases.

\subsection{Decoder}
The goal of the decoder network is to reconstruct a given input as accurately as possible, given the latent code produced by the encoder. 

\paragraph{Tree decoding} We use the same tree structure for decoding as we used for encoding. . Additionally, \cite{fabius2015variational} shows that having a reversed order of the input sequence compared to the reconstructed sequence when dealing with autoregressive models improves the performance. We employ this technique in our model, which means that since our encoder processes trees bottom-up, the decoder will produce trees top-down. The idea here is that the first steps of decoding the tree are more related to the latent space than the last steps.

A method called the doubly-recurrent neural network (DRNN) \cite{alvarezmelis2017tree} allows for top-down tree generation from an encoded vector representation. This method operates solely on the vector representation and does not require that either the tree structure or the nodes are given. The DRNN is based on two recurrent neural networks, breadth and depth-wise, to model ancestral and fraternal information flow. Hence, each node in the tree receives input from two types of states, one input comes from its parent, and the other comes from its previous sibling. Since we capture the information flow with two separate RNN modules, we obtain two hidden states for each node in the tree (ancestral and fraternal). These hidden states can be combined to form a so-called predictive hidden state from which topological information and node labels can be predicted.

For some node $y$ with parent $pa(y)$ and previous sibling $s(y)$, the ancestral and fraternal hidden states are computed as follows:

\begin{align}
    \mathbf{h}_a^y &= rnn_a(\mathbf{h}_a^{pa(y)}, \mathbf{i}^{pa(y)}) \\ \label{eq:ancestral_update}
    \mathbf{h}_f^y &= rnn_f(\mathbf{h}_f^{s(y)}, \mathbf{i}^{s(y)}) 
\end{align}

Where $rnn_a$, $rnn_f$ are functions that apply one step of the ancestral and fraternal RNNs, respectively. Furthermore, $\mathbf{i}^{pa(y)}$, $\mathbf{i}^{s(y)}$ are the input values (label vectors) of the parent and previous sibling respectively. After the ancestral and fraternal states of $y$ have been computed with the observed labels of its parent and previous sibling, these states can be combined to form a predictive hidden state:

\begin{align}
    \mathbf{h}^y_{pred} = \tanh\left((\mathbf{W}_a \cdot \mathbf{h}_a^y + \mathbf{b}_a) + (\mathbf{W}_f \cdot \mathbf{h}_f^y + \mathbf{b}_f)\right)
\end{align}

Where the operations applied to $\mathbf{h}_a^y$, $\mathbf{h}_f^y$ are linear layers with learnable weights and biases. This combined state then contains information about the nodes' surroundings in the tree.

For each node in the tree, the model needs to decide whether it has offspring and whether it has any successor siblings. Answering this question for every node allows the model to construct a complete tree from scratch. This method avoids using terminal tokens to end the generation compared to sequential decoders. In turn, this method allows us to make topological decisions explicitly bypassing the need for padding token. We can use the predictive hidden state of a node $\mathbf{h}^y_{pred}$, with a linear layer and a sigmoid activation to compute the probability for offspring and successor siblings as:

\begin{align}
    p_a^y &= \sigma(\mathbf{W}_{pa} \cdot \mathbf{h}_{pred}^y + \mathbf{b}_{pa}) \label{eq:prob_ancestral} \\
    p_f^y &= \sigma(\mathbf{W}_{pf} \cdot \mathbf{h}_{pred}^y + \mathbf{b}_{pf})\label{eq:prob_fraternal}
\end{align}

Where during training, we use the actual values for whether a node has children and successor siblings. During inference, we can either greedily choose any confidence level to continue creating offspring and succeeding siblings by checking whether the probability is above some threshold or sample this choice.

Besides topological predictions, the model should also predict the label of each token. Again the predictive hidden state can be used for label prediction as follows:

\begin{align}
    \mathbf{o}^y =  softmax\left(\mathbf{W}_o \cdot \mathbf{h}_{pred}^y + \mathbf{b}_{o}\right) \label{eq:label_pred}
\end{align}

\paragraph{Tree decoding optimizations} Now that we have the basic DRNN model \cite{alvarezmelis2017tree} in place to generate a tree from scratch using a latent vector, we can optimize it for our use case. There are still a few issues with the tree decoding method, such that it is not practical to be used with generating industry-grade programming languages.

The first issue is the possibly infinitely large vocabulary that source code allows. In contrast to NLP, developers may choose any combination of characters, e.g., for identifiers, and are not limited to a finite dictionary of words. For identifiers, the compiler does not care about the identifier itself but only about the declaration of an identifier and its references. Changing these identifiers to any other random combination of characters while preserving the relations will not functionally change the code. Therefore, we map each unique identifier to a reusable ID \cite{tufano2019learning} and treat the prediction of identifiers as a clustering problem between declarations and references. We use the predictive hidden states of the nodes to learn relationships between declarations and references.

The model can keep track of a list of the declared identifiers while generating an AST. Each time a new identifier is declared, a new reusable ID is added to the list. Then for each reference, we can compute the similarity to each of the declared identifiers using some similarity function and predict the most similar identifier. We can keep track of what type of node we are currently trying to predict due to the AST structure and because we have access to the parent node label, i.e., the parent node indicates whether the child node is a declaration or reference. Let $D$ be the set of currently declared identifier nodes and $y$ be the current reference node we are trying to predict, the most similar declared identifier can be computed as follows:

\begin{align}
    \mathbf{s}^{yz} &= similarity(\mathbf{W}_c \cdot \mathbf{h}^y_{pred} + \mathbf{b}_c,  \mathbf{W}_c \cdot \mathbf{h}^z_{pred} + \mathbf{b}_c) \\
    \mathbf{r}^y &= \min_{z \in D}(\mathbf{s}^{yz})
\end{align}

We have a similar problem for literal tokens; developers can use an almost infinitely large number of unique literals in source code. However, in contrast to identifier tokens, literal tokens influence the functionality of a program. Therefore, to assure that generated programs are still compile-able, we cannot remap the literal tokens to reduce the token count. For example, we cannot map rarely used literals to special unknown tokens, as unknown tokens create compiler errors. Instead, we can employ adaptive softmax \cite{grave2017efficient} to use a vocabulary consisting of many unique literal tokens without a considerable increase in computational complexity.

We have identifiers and literals as token categories already, but we can also categorize the leftover tokens into the following categories:

\begin{itemize}
    \item \textbf{Reserved tokens:} for, if, while, ...
    \item \textbf{Types:} int, long, string, ...
    \item \textbf{Built in function names:} printf, scanf, push, ...
\end{itemize}

In total, the five categories cover all the different tokens of the programming language (at least for C++). The reason for splitting up the leftover tokens into more categories is to predict these categories separately based on their parent node. For example, this ensures that we do not input a type-token in the tree, where there should be a reserved token. The categorization improves the compilation rate of the generated programs by allowing the model only to predict tokens of the correct token category. The tree-structured representation during decoding allows us to use this optimization technique. For the reserved tokens, type, and built-in function names, \cref{eq:label_pred} is used for label prediction, as there is only a limited number of unique tokens in these categories.

To allow for the categorized label predictions, we need to add one more element to the DRNN model. As mentioned, the chosen category to predict a label for depends mainly on the parent node. An essential aspect of the tree structure is that all categories, except for reserved tokens, only occur on leaf nodes. In essence, all types, identifiers, built-in function names, and literals occur in the leaves of the trees. Therefore, if a node has offspring, the category of the current node must be a reserved token. However, if a node has no offspring, it can be either of the categories, and we need to somehow decide which category to predict a label for. Note that a reserved token can also be on a leaf node on the tree. For example, consider an empty return statement. For that reason, similar to the topology predictions, we have the model predict whether a node is of the reserved token category or not. This prediction  is computed in the same way as the topology predictions using the predictive hidden state of the node as follows: 

\begin{align}
    p_r^y &= \sigma(\mathbf{W}_{pr} \cdot \mathbf{h}_{pred}^y + \mathbf{b}_{pr}) \label{eq:res_pred}
\end{align}

Next, let us consider the case where a node is not of the reserved token category. The model must be informed to predict a label for one of the remaining categories. The data in the tree structure can be represented in a way that this choice depends on the parent label. We can label the parent node of literals: `literal', type nodes: `type', and built-in function names: `built-in function name'. The identifier (including references) token category is the leftover category that can be predicted if a node does not fall in any of the previous categories.

\begin{figure}[ht!]
    \centering
    \includegraphics[width=\linewidth]{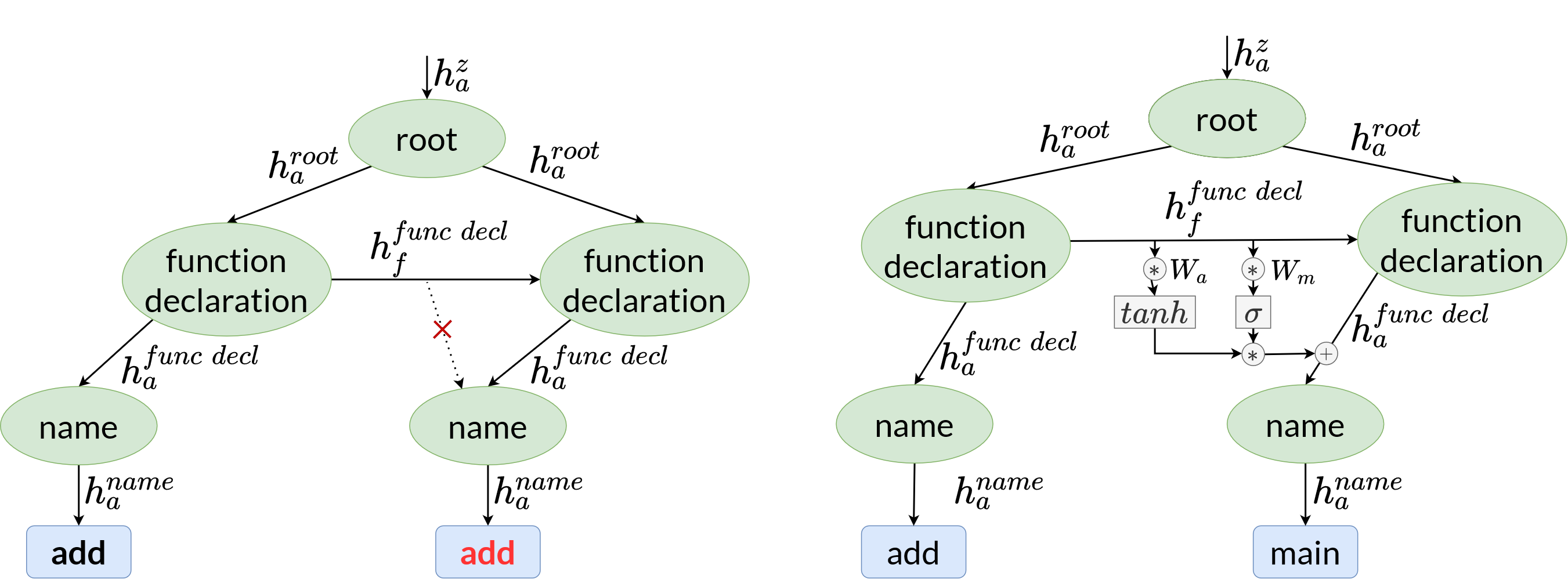}
    \caption{DRNN expanded with an add gate to allow for information flow from previous siblings downwards}
    \label{fig:treeAddGate}
\end{figure}

\paragraph{Add gate} The DRNN model has a large flaw, where it is not able to differentiate between paths with the same prefix. For example, consider the situation depicted in \cref{fig:treeAddGate}, where we have two function declarations named `add' and `main'. Due to the information flow downwards, both name nodes have the same hidden state and the model is not able to distinguish the leaf nodes and will therefore predict the same label for both. This issue is depicted in the left image of \cref{fig:treeAddGate}. To solve this issue, we would like to incorporate the fraternal states in the downwards flow for the model to learn to differentiate the paths downwards. Hence we would like to revise \cref{eq:ancestral_update}, where we take inspiration from the LSTM model and apply the idea of the add gate to our ancestral update formula as follows:

\begin{align}
    &\mathbf{m}_f^y = \sigma(\mathbf{W}_m \cdot \mathbf{h}_f^y + \mathbf{b}_m)\\
    &\mathbf{a}_f^y = tanh(\mathbf{W}_a \cdot \mathbf{h}_f^y + \mathbf{b}_a)\\
    &\mathbf{h}_a^y = \mathbf{h}_a^y + (\mathbf{a}_f * \mathbf{m}_f)
\end{align}

This update to the fraternal state is applied after predicting the label for node $y$, which is depicted in the right image of \cref{fig:treeAddGate}. Here, $a_f^y$ is the value of the transformation on the previous sibling state that should be added to the parent state, where the $tanh$ transforms it between -1 and +1 to mitigate exploding gradients. Furthermore, $m_f^y$ decides which elements should be added using a sigmoid function that outputs values between 0 and 1. By multiplying $a_f^y$ with $m_f^y$, the model can learn to decide what and how much to add from the previous sibling state to each parent state's element to help predict the next steps of the tree.

\subsection{Optimization}

\subsubsection{Mitigating KL vanishing}
KL vanishing is a common issue when dealing with VAEs with a decoder parameterized by an auto-regressive model.
We mitigate it vanishing using cyclical KL cost annealing \cite{fu2019cyclical}. Furthermore, we apply pooling to the hidden states of the RNN network in the encoder. Long \textit{et al.} \cite{long2019preventing} show this pooling method can effectively prevent the posterior collapse issue. The argumentation is that taking only the last hidden state from the RNN of the encoder will cause the encoder to often produce similar representations leading to nearly indistinguishable samples. In turn, the decoder will ignore the latent variables as they convey no useful information about the data. Pooling the hidden states of the encoder RNN incorporates more information of the entire input into the latent variables mitigating the KL vanishing issue.

\subsubsection{Loss function}
We can categorize two classes of predictions that the model has to perform to generate a tree structure: labels and topology. The topology class entails the predictions that lead to building the structure of the tree. Precisely, for each node, predict whether there is offspring and any successor siblings. The label class encompasses filling in the actual values in the tree structure, predicting what tokens should occur at a particular node in the tree given the surrounding context.

Let us first look into the loss terms for the topology predictions. Predicting whether a node has offspring and successor siblings are binary choices. Recall from \cref{eq:prob_ancestral} and \cref{eq:prob_fraternal} that we compute probabilities for these choices using the sigmoid function that gives values between 0 and 1. Therefore, we can use binary cross-entropy to compute the loss for both of the topology predictions. The loss can be computed between the predicted probabilities and the true values of having offspring and successor siblings. Let $a^y$, $f^y$ represent the actual values of having offspring and successor siblings for node $y$, the topological losses for this node are then computed as follows:

\begin{align}
    \mathcal{L}_{a}(y) = - a^y \cdot \log(p^y_a) + (1 - a^y) \cdot \log(1 - p^y_a) \\
    \mathcal{L}_{f}(y) = - f^y \cdot \log(p^y_f) + (1 - f^y) \cdot \log(1 - p^y_f)
\end{align}

\noindent where $\mathcal{L}_{a}$, $\mathcal{L}_{f}$ denote the ancestral and fraternal loss respectively. Because the reserved token category prediction (\cref{eq:res_pred}) is so similar to the topological predictions, the loss for that component can be defined in a similar fashion:

\begin{align}
    \mathcal{L}_{r}(y) = - r^y \cdot \log(p^y_r) + (1 - r^y) \cdot \log(1 - p^y_r)
\end{align}

\noindent Where we define $r^y$ to represent the actual value of node $y$ being in the reserved token category.

Secondly, let us elaborate on the loss for the label predictions. For all label categories, except the identifiers, we are dealing with a classification problem. Hence, we can compute the cross entropy loss (or negative log likelihood) between the softmax output for a node $y$ (\cref{eq:label_pred}) and its true label:

\begin{align}
    \mathcal{L}_{l}(y) = - \log(\mathbf{o}^y[l^y])
\end{align}

\noindent Where we assume that $l^y$ is the index of the true label, and hence $\mathbf{o}^y[l^y]$ retrieves the softmax value at the index of the correct class. Lastly, since predicting the labels of identifier (reference) tokens is treated as a clustering problem, we can use triplet loss \cite{chechik2010large}. We use a similarity function (or inversely, distance function) and maximize the similarity between declarations and reference labels. To compute the loss of a reference node $y$, we select the true declaration node $z$ and sample a negative declaration node $x$; the loss is then defined as follows:

\begin{align}
    \mathcal{L}_i(y)=\max(\mathbf{s}^{yx} - \mathbf{s}^{yz},0)
\end{align}

\noindent We can then combine all of the separate components to form a single reconstruction loss function for a node:

\begin{small}
\begin{align}
   \mathcal{L}_{rec}(y) = 
\begin{cases}
    \mathcal{L}_{a}(y) + \mathcal{L}_{f}(y) + \mathcal{L}_{r}(y),& \text{if } y \text{ is a declaration} \\
    \mathcal{L}_{a}(y) + \mathcal{L}_{f}(y) + \mathcal{L}_{r}(y) + \mathcal{L}_{i}(y),& \text{if } y \text{ is a reference} \\
    \mathcal{L}_{a}(y) + \mathcal{L}_{f}(y) + \mathcal{L}_{r}(y) + \mathcal{L}_{l}(y),& \text{otherwise}
\end{cases}
\end{align}
\end{small}

\noindent Because the loss is decoupled, this allows us to weigh the objectives differently to emphasize, for example, topology or label prediction accuracy. We leave experimenting with different weights for objectives as future work. Let $N$ be the set of nodes of a tree; we can define the loss reconstruction loss of an entire tree as follows:

\begin{align}
    \mathcal{L}_{tot\_rec}(N) = \sum_{y \in N}\mathcal{L}_{rec}(y)
\end{align}

\noindent The total loss function, combining the KL divergence, KL weight $w$ and reconstruction loss becomes:

\begin{align}
    \mathcal{L}(N) = \mathcal{L}_{tot\_rec}(N) - w \cdot D_{KL}\left(Q(z|N)||P(N)\right)
\end{align}

\noindent During training, we perform teacher forcing, technique that is commonly used with sequence generation.

\begin{figure*}[ht!]
    \centering
    \includegraphics[width=\linewidth]{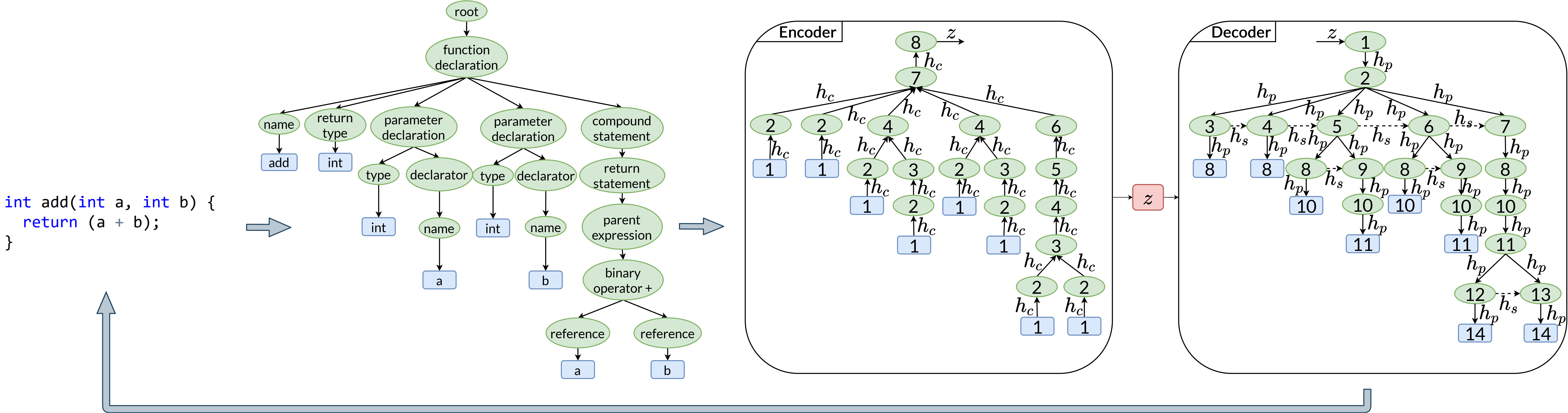}
    \caption[Tree2Tree model high-level overview]{Tree to tree autoencoder overview. \textbf{First Fig.}: The piece of code considered. \textbf{Second Fig.}: The piece of code parsed to an AST tree format. \textbf{Third Fig.}: The order in which the encoder module encodes the tree structure bottom-up. Here, $h_c$ indicates the hidden state that travels from a child to a parent. \textbf{Fourth Fig.}: The order in which the decoder module decodes the tree structure top-down. Here, $h_p$ indicates the hidden state that travels from a parent to a child, and $h_s$ indicates the hidden state that travels from a node to its successor sibling.}
    \label{fig:tree2treeVAE}
\end{figure*}

\subsection{Data Pre-processing}
Now that we have defined the proposed Tree2Tree model, we have to pre-process the C++ programs to the representation that fits the model's description. We have to convert the sequential source code to the hierarchical tree structure. In this section, we will first go over the process of transforming the data to the tree structure and back. Afterward, we will discuss the filters we apply to the data.

\subsubsection{Transformations}
The first steps are to remove comments, extract imports and expand macros. The resulting output is the actual source code. Next, we extract the AST representation from the Clang C++ compiler. We filter the extracted AST to the minimal representation required to reconstruct the source code. Moreover, we categorize each node in the tree in one of the following categories: reserved tokens, types, built-in function names, literals, and identifiers. An example of how a piece of code is transformed into a tree structure is depicted in \cref{fig:tree2treeVAE}. To go back to the sequence representation from the tree representation, we start from the root and recursively iterate over the children to reconstruct the source code.

\subsubsection{Filtering}
To keep the training stage of the model efficient, we set a limit to the maximum tree size the programs may have. Large programs will create a bottleneck during training, especially with mini-batch processing. We find that putting the tree size limit at 750 strikes a good balance between effective training and keeping a large portion of the data because 95\% of the programs are within this limit of 750 tree nodes. Therefore, we filter out any programs with a tree size of more than 750 nodes and use the remaining data set for training and validating the model. In the next section, we introduce a baseline model where we apply the same filters on the data as with the Tree2Tree model, to allow for a fair comparison of the two models.

%% file: Experiments.tex
This section provides a set of experiments to analyze how accurately the proposed autoencoder can reconstruct and generate programs. First, we describe our evaluation dataset. Next, we discuss the implementation details of our model, then we introduce a baseline to compare our model to, and finally, we show the evaluation results. 

\subsection{Dataset}
We train and evaluate our model on a dataset of programs from code competition websites. Programs from these platforms exhibit a few qualities that are suitable for program synthesis. The programs are tested and known to be syntactically correct and compile-able, and they are standalone code fragments and do not depend on any code that is not built into the programming language. The dataset consists of almost two million C++ programs across 148 competitions divided over 904 problems. 


\subsection{Implementation details}
We use three-layered LSTMs in the encoder and decoder with a recurrent dropout rate of 20\% to reduce over-fitting. The embedding layer is initialized with Glove wiki gigaword 50 \cite{pennington2014glove} embedding. We train the model using the Adam optimizer \cite{kingma2014adam} with a learning rate of $1e-3$ and 10 epochs with early stopping and a patience of 3. We train and run the experiments on GPUs with a batch size set to 32.

\subsection{Baseline}
\begin{figure*}[ht!]
    \centering
    \includegraphics[width=0.8\linewidth]{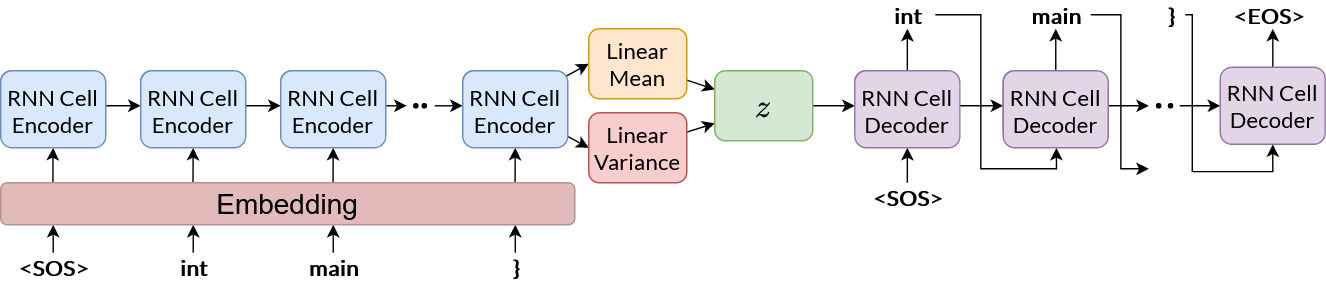}
    \caption{The architecture of the Seq2Seq model}
    \label{fig:seq2seq}
\end{figure*}

Our method is compared to a baseline inspired by autoencoders used for text generation in natural language. We can also evaluate how well these models generalize to source code synthesis by taking inspiration from natural language models. The model architecture is inspired from \cite{bowman2015generating}. In this architecture, both the encoder and decoder networks contain single-layer recurrent neural networks. A Gaussian prior is used for the regularization of the latent space. The model operates on the original sequences of source code and decodes the latent vector back to the source code without an intermediate structured representation. Therefore we refer to the baseline model as the Sequence-to-Sequence (Seq2Seq) model, and the architecture is depicted in \cref{fig:seq2seq}.

Similar to our proposed autoencoder model, we employ methods to mitigate KL-vanishing. Again, we use cyclic KL annealing \cite{fu2019cyclical}, and we combine this with a technique called word dropout \cite{bowman2015generating} to weaken the decoder.

\subsection{Reconstruction results}\label{results:recon}
First of all, we look at how accurately the autoencoders can reconstruct programs. We use a separate test split containing around 60.000 samples of our data set to evaluate this and use these samples as input for the autoencoders.

We compute BLEU scores \cite{papineni2002bleu} for both models on the original representation of the source code to obtain comparable results, i.e., we do not use the tree representation. The Tree2Tree model thus has an extra step to use the data parser to transform the tree representation back to source code. This extra step is disadvantageous for the Tree2Tree model as it may introduce some errors due to imperfections in the parsing process. The BLEU scores are then computed on each token in a program: keywords, identifiers, operators, and special symbols such as semicolons or braces. We report on the cumulative BLEU-1 through BLEU-4 scores to indicate the overlap between original and reconstructed programs. Furthermore, we present the percentage of reconstructed programs that compile to indicate how well the models have learned the programming language's syntax. We experiment with different combinations of latent sizes $l$ and hidden RNN dimensions $h$:  ($l$:10,  $h$:50), ($l$:50, $h$:100), ($l$:100, $h$:200), ($l$:150, $h$:300), ($l$:300, $h$:500), ($l$:500, $h$:800), ($l$:800, $h$:1200).

For reconstructions, we will use greedy decoding in both models. That means that for each label, we select the most likely prediction according to the model. We take this approach as we would like the reconstructions to be close to the original input. In contrast, sampling will give more variety in the output, which can be helpful when generating new samples. The results of the reconstruction experiments using greedy decoding are shown in \cref{tab:rec_results}.

\begin{table}[ht!]
\centering
\begingroup
\setlength{\tabcolsep}{3pt} 
\renewcommand{\arraystretch}{1.4} 
\begin{tabular}{cccccccc}
\ChangeRT{1pt}
\textbf{Model}   & \textbf{Latent size} & \textbf{BLEU-1} & \textbf{BLEU-2} & \textbf{BLEU-3} & \textbf{BLEU-4} & \textbf{Compiles}\\ \hline
\multirow{7}{*}{Seq2Seq}    &   10   &  0.037   &    0.024     &     0.017     &    0.013       &    0.000\%   \\
                            &   50   &      0.085    &      0.061         &         0.047      &    0.037       &       42.467\%      \\
                            &   100   &  0.295   &      0.225     &     0.176      &    0.141       &   65.808\%               \\
                            &   150   &     0.278  & 0.211          & 0.165          & 0.131          &  66.971\%              \\
                            &   300   & 0.346                     &  0.262                        &     0.203                     &     0.161                     & 60.651\% \\   
                            &   500   &  0.421                   &      0.332                    &         0.263                 &      0.211                    &  90.329\% \\
                            &   800   &  0.429                   &      0.329                    &         0.253                 &      0.195                    &  \textbf{91.784\%} \\\hline
\multirow{7}{*}{Tree2Tree}  &   10   &  0.445  &    0.339     &     0.260     &     0.202      &   28.375\%    \\
                        &   50   &     0.417    &      0.317    &       0.242    & 0.189  &  23.256\% \\
                            &   100   &     0.423      &      0.323     &     0.251      &  0.200 &      30.429\%     \\
                            &   150   & \textbf{0.486} &     \textbf{0.382}    &  \textbf{0.302}     &      \textbf{0.243}    &   35.419\%           \\
                            &   300   &   0.457                 &     0.342                &      0.260                &     0.202                   & 35.054\% \\   
                            &   500   &     0.398                &      0.301             & 0.230      &             0.178             &            36.022\%      \\
                             &   800   &     0.258                &      0.182             & 0.131      &             0.096             &     2.358\%      \\\ChangeRT{1pt}
\end{tabular}
\endgroup
\caption{Reconstruction results.}
\label{tab:rec_results}
\end{table}

The results listed in \cref{tab:rec_results} show the superiority of the Tree2Tree model in terms of reconstruction capability (BLEU scores), especially for smaller latent sizes. The reconstruction scores of the Tree2Tree model of latent size 150 outperform all the Seq2Seq models up to latent size 800. In contrast, the Seq2Seq models show to perform much better at constructing compile-able programs, which improves with the model's size, to nearly 100\%. This is a surprising result, which is investigated in more detail in supplementary material\footnote{\url{https://tree2tree.app/supmat.pdf}}.

An interesting result is that the performance of the models does not necessarily increase with the size of the model. Especially for the Tree2Tree models, we see that after latent size 150, the models' performance decreases. In general, one would expect that the model would perform better with an increase in latent size, allowing more information flow between the encoder and decoder. We hypothesize that, because not only the latent size increases but also the number of hidden units in the auto-regressive models, the models experience KL vanishing. Due to the increasing hidden units, the auto-regressive models become stronger and may depend more on their predictions, ignoring information from the latent vector. In turn, the reconstruction performance vastly decreases. Confirmation of this hypothesis is left as a venue for future work.

Next, we would like to experiment on how different input sizes affect the performance of both models. Due to the tree-structured representation used by the Tree2Tree model, the size of the sequences that the RNNs process scale proportionally to the width and depth of the tree. The Seq2Seq model, on the other hand, processes sequences left to right, hence the number of computations of the RNNs scale directly with the sequence length.

To evaluate the performance on different sized inputs, we split the test data set into three subsets. A small, medium and large subset with the following properties:

\begin{itemize}
    \item \textbf{small subset}: maximum of 250 tokens
    \item \textbf{medium subset}: between 251 and 500 tokens
    \item \textbf{large subset}: between 501 and 750 tokens
\end{itemize}

We compute the BLEU scores and compilation percentage again using greedy decoding on the smaller subsets for the best performing Seq2Seq and Tree2Tree models, based on the results of \cref{tab:rec_results}. Here, performance is based on the combination of BLEU-4 and compilation percentage. For Seq2Seq, this is the model with latent size 500. Similarly, for Tree2Tree, this is the model with latent size 150. The results are depicted in \cref{tab:rec_results_inp_sizes}.



\begin{table}[ht!]
\centering
\begingroup
\setlength{\tabcolsep}{3pt} 
\renewcommand{\arraystretch}{1.4} 
\begin{tabular}{clccccc}
\ChangeRT{1pt}
\textbf{Model}   & \textbf{Input size} & \textbf{BLEU-1} & \textbf{BLEU-2} & \textbf{BLEU-3} & \textbf{BLEU-4} & \textbf{Compiles}\\ \hline
\multirow{3}{*}{Seq2Seq}    &   small   &   0.513  &    0.403    &      0.321    &  0.258      &  95.334\%     \\
                            &   medium   &  0.306      &    0.244        &      0.192     & 0.153      & 86.812\% \\
                            &   large   &    0.196   &      0.157   &   0.123       &   0.096       &  87.971\% \\ \hline
\multirow{3}{*}{Tree2Tree}  &   small   &   0.633  &    0.516     &     0.424     &     0.355      &  59.022\%                \\
                            &   medium   &   0.478  &   0.371        &          0.289 &     0.229   & 21.241\%            \\
                            &   large   &  0.324 &      0.242  & 0.181    &    0.138     &    5.001\%         \\ \ChangeRT{1pt}
\end{tabular}
\endgroup
\caption{Reconstruction results of the best models on different input sizes.}
\label{tab:rec_results_inp_sizes}
\end{table}

From \cref{tab:rec_results_inp_sizes} we can observe that both models follow the same logical trend: the larger the input size, the lower BLEU-scores and compilation percentages. For the Tree2Tree model, the BLEU scores for the medium subset seem to be similar to the BLEU scores on the entire test set, whereas, for the Seq2Seq model, the BLEU scores are much lower on the medium subset. The models seem to be fairly close in terms of performance degradation from small to large program sizes. For example, we can measure performance degradation for the large versus small subset by dividing the BLEU-4 scores on the large set by the BLEU-4 score on the small set. For the Seq2Seq model, we get a score of $0.372$, and for the Tree2Tree model, we get $0.389$. Similarly, we get $0.593$ and $0.645$ for the Seq2Seq and Tree2Tree model for the medium versus small subset. While the performance degrades less with increasing input sizes for the Tree2Tree model, this difference is insignificant.

An issue with the aforementioned computation of performance degradation is that it does not correct for elements in programs that are almost always present. For example, each program contains a main function, with standard input and output streams. The models may simply always predict these standard elements of a program and then use the information of the encoder to complete the details of the program. However, this causes the BLEU score to consist of two parts: the score for the prediction of the elements that are always present and the score of what it has learned to predict together with the encoder. The latter is more interesting and shows how much information can be saved in the latent vector.

Therefore, we apply a correction on the BLEU scores to focus on the prediction based on the information in the latent vector. We compute corrected scores by feeding the decoder with random latent vectors and computing BLEU scores on the subsets of the test data set. Then, we subtract these correction scores from the computed BLEU scores in \cref{tab:rec_results_inp_sizes}, and take 0 if the result of the subtraction is negative. The corrected BLEU scores including the correction scores are presented in \cref{tab:corrected_rec_results_inp_sizes}.

\begin{table}[ht!]
\centering
\begingroup
\setlength{\tabcolsep}{3pt} 
\renewcommand{\arraystretch}{1.4} 
\resizebox{\linewidth}{!}{%

\begin{tabular}{clccccc}
\ChangeRT{1pt}
\textbf{Model}   & \textbf{Input size} & \textbf{BLEU-1} & \textbf{BLEU-2} & \textbf{BLEU-3} & \textbf{BLEU-4} \\ \hline
\multirow{3}{*}{Seq2Seq}    &   small   &   0.072 (0.441)  &    0.077 (0.326)    &  0.075    (0.246)    &  0.070 (0.188)     \\
                            &   medium   & 0.006 (0.300)      &  0.018  (0.226)        &    0.021  (0.171)     & 0.023 (0.130)    \\
                            &   large   &   0.000 (0.213)   &  0.000    (0.166)   & 0.000  (0.128)       &  0.000 (0.099)   \\ \hline
\multirow{3}{*}{Tree2Tree}  &   small   &  0.200 (0.433)  &  0.220  (0.296)     &  0.223   (0.201)     &    0.218 (0.137)        \\
                            &   medium   &  0.148 (0.330)  &  0.147 (0.224)        & 0.146 (0.150) &    0.128 (0.101)        \\
                            &   large   &  0.102 (0.222) &    0.090  (0.152)  &  0.079 (0.102)     &  0.070  (0.068)      &   \\ \ChangeRT{1pt}
\end{tabular}%
}
\endgroup
\caption{Corrected BLEU scores of reconstructed results of the best models on different input sizes. (correction scores in parenthesis)}
\label{tab:corrected_rec_results_inp_sizes}
\end{table}

\Cref{tab:corrected_rec_results_inp_sizes} indicates a large difference in performance degradation between the Seq2Seq model and the Tree2Tree model. A noticeable result is that the corrected BLEU scores for large programs predicted by the Seq2Seq model are 0. Hence, the Seq2Seq model extracts no information from the latent vector at all for large programs. Similarly, for medium-sized programs, little information is transferred between the encoder and decoder. We can again compute the performance degradation scores for the Seq2Seq model, which are $0.280$ and $0.00$ for the medium versus small and large versus small subsets, respectively, on the corrected BLEU-4.

In contrast, the performance degradation is much smaller for the Tree2Tree model:  $0.587$ and $0.321$ for the medium versus small and large versus small subsets, respectively, on the corrected BLEU-4. Hence, the structural nature of the Tree2Tree model scales better to large input sequences than the Seq2Seq model in terms of reconstruction scores, even with a much smaller latent size. We hypothesize that, due to the Tree2Tree model performing auto-regressive operations on paths of trees that scale on the width and depth of the tree, the model mitigates exploding and vanishing gradients.

An interesting observation is that the latent vector conveys relatively little information in terms of BLEU scores. The correction scores make up a large part of the total BLEU scores as presented in \cref{tab:rec_results_inp_sizes}. Hence, the BLEU scores are largely determined by the models' general knowledge of how C++ programs are built up and not the specific content.  This is a side effect of training on this particular data set, as the model will put a lot of its focus on learning the highest reconstruction score, which can be obtained by learning these basic constructs of a program.

\subsection{Generative results}
The aim of the autoencoders is to be able to generate syntactically correct and compile-able programs from latent space. Therefore, we evaluate both the autoencoders on this ability next. To do this, we sample 1000 random latent vectors from the prior distribution $\mathcal{N}(0, I)$ and input these vectors to the decoder networks. Then, we compute the percentage of generated programs that compiles and is thus also syntactically correct. This measure allows us to see how well autoencoders can generate reasonable samples from any point in latent space that conform to the C++ syntax. Again, we experiment with different combinations of latent sizes and hidden dimensions.

We employ two decoding strategies to test the generative capabilities of the models: greedy decoding and sampling. We use greedy decoding to see how well the models perform when selecting their most likely predictions. Sampling is more often used when generating new samples, allowing for more diversity in the output. The sampling strategy we apply is a combination of top-$k$, nucleus, and temperature sampling \cite{holtzman2019curious}. We first use temperature sampling to scale the logits to control the shape of the probability distribution. Then, we filter the on the top-$k$ samples, after which we filter tokens on their cumulative probability using nucleus sampling (top-$p$). Lastly, we sample a token from the resulting distribution. The selected sampling hyper-parameters for this experiment are: $k=40$, $p=0.9$, $temperature=0.7$. The results of the experiment are displayed in \cref{tab:gen_results}.

\begin{table}[ht!]
\centering
\begingroup
\setlength{\tabcolsep}{3pt} 
\renewcommand{\arraystretch}{1.4} 
\begin{tabular}{cccc}
\ChangeRT{1pt}
\textbf{Model}   & \textbf{Latent size} & \textbf{Greedy search} & \textbf{Sampling} \\ \hline
\multirow{5}{*}{Seq2Seq}    &   10   &  0.0\%   & 0.9\%  \\
                            &   50   &   38.5\%    &  2.9\%   \\
                            &   100   &   62.1\%   &  21.3\%     \\
                            &   150   &   58.0\%  &     23.5\%        \\
                            &   300   &  60.6\%  &  36.8\%   \\   
                            &   500   &  67.5\% & 37.8\% \\
                            &   800   & 78.2\%  & 39.6\% \\\hline
\multirow{5}{*}{Tree2Tree}  &   10   &   29.6\%   & 20.2\% \\
                            &   50    & 22.6\%   &  17.7\%    \\
                            &   100    & 30.3\%  &       22.1\%      \\
                            &   150  &  26.9\%  &   18.8\%     \\
                            &   300  & 23.4\%  & 12.8\%    \\   
                            &   500 & 25.6\%  & 14.4\%\\
                            &   800   &  4.1\% & 6.7\% \\\ChangeRT{1pt}
\end{tabular}
\endgroup
\caption{Generative results compilation percentage.}
\label{tab:gen_results}
\end{table}

The results from \cref{tab:gen_results} show similar trends as \cref{results:recon}. The general trend is: the larger the model (in terms of latent size and hidden units), the higher the compilation percentage. Moreover, greedy search during inference gives a higher compilation percentage than sampling. This outcome is not surprising, as, with greedy search, we always pick the label for which the model is most confident. On the other hand, sampling gives a more varied output and may be useful for searching similar programs in a vicinity of the latent space. The trade-off for a more diverse output is thus a lower compilation ratio. 

\paragraph{General errors} We have manually inspected the messages of the compiler to find the main issues with the generated programs. This shows the weaknesses of the current programs and what a possible improved model can focus on to improve the performance of the compilation rate. The most common issues with the Seq2Seq model are:
\begin{itemize}
    \item References to identifiers not declared in scope
    \item Re-declarations of identifiers
    
\end{itemize}

Furthermore, we have also inspected the compiler errors on the generated programs of the Tree2Tree models, the most often occurring errors are:
\begin{itemize}
    \item References to identifiers not declared in scope
    \item Invalid types  (e.g. trying to access an array element on a non-array type)
    \item Mismatched types (e.g. concatenate integer and string)
    \item Invalid type conversion (e.g. read standard input to an integer array)
    \item Use of member functions not available for a type (e.g. try to use the $push\_back$ member function of integer arrays on a string) 
\end{itemize}

Lastly, we have performed a more qualitative evaluation (see supplementary material, \url{https://tree2tree.app/supmat.pdf}). From this evaluation, we find that the Tree2Tree model has a structured latent space where similar points in latent space also map to similar programs. This allows for a directed search over the latent space. Moreover, we find that the Seq2Seq model maps the same program to multiple latent vectors, indicating some form of KL vanishing.

%% file: conclusion.tex
Our experimental results on the code competition data set show that our proposed tree-structured model significantly outperforms the baseline model based on reconstruction performance. Furthermore, we find that our proposed model scales considerably better to larger input sizes than the baseline model. Our model can reconstruct compile-able programs up to at least a size of 750 tokens.

Due to the structural nature of our model, the auto-regressive operations are performed on paths of trees instead of linear sequences as with the baseline model. Therefore, the size of the sequences that our model processes scales to the width and depth of a tree instead of the program's linear size. We hypothesize that this mitigates the common problem of exploding and vanishing gradients, which causes our model to degrade much less in performance with increasing input size compared to the baseline model.

Moreover, our proposed model has learned a structured latent space where coordinates close in the latent space decode into similar programs. This is a valuable property for GP, as this allows for a directed search over the latent space.

Additionally, we showed that greedy search during inference results in a higher percentage of compile-able programs than sampling. This introduces a trade-off where sampling may yield more diverse programs while fewer programs are syntactically correct and compile-able. Furthermore, our proposed autoencoder model can generate syntactically correct and compile-able programs from latent space with a compilation rate of over 25\%.

Lastly, our experimental results showed that while the baseline model achieved a much higher compilation ratio, this is mainly caused by KL vanishing. It is difficult to mitigate the KL vanishing in the baseline model, even with special techniques to alleviate this issue. Therefore, the model tends to map multiple latent vectors to the same programs. The baseline model thus maps all latent vectors to a pool of programs it remembers. This pool of programs tends to have a high compilation percentage. However, an autoencoder that can only generate a small set of programs may not be useful for program synthesis.

A limitation of our work is that the Tree2Tree model currently predicts references to identifiers from a list of all identifiers defined before in the program. However, this method does not keep track of the scope of a program. Consequently, the model predicts references to identifiers that are not in scope at that point of the program, leading to undeclared references. An interesting future direction is to keep track of the scope of the current point in the program to only refer to identifiers that are declared in the current scope, which could improve the number of syntactically correct and compile-able programs generated by the model.

Further interesting future directions are: evaluating the proposed model architecture on different programming languages, experimenting with combinations of sequence and tree architectures, considering transformers or CNNs instead of only RNNs, and experimenting with different data sets that more accurately represents the complete taxonomy of C++ programs.

%% file: bare_conf.bbl
\begin{thebibliography}{10}

\bibitem{alon2019structural}
U.~Alon, R.~Sadaka, O.~Levy, and E.~Yahav.
\newblock Structural language models for any-code generation.
\newblock {\em arXiv preprint arXiv:1910.00577}, 2019.

\bibitem{alvarezmelis2017tree}
D.~Alvarez-Melis and T.~S. Jaakkola.
\newblock Tree-structured decoding with doubly-recurrent neural networks.
\newblock In {\em Proceedings of the International Conference on Learning
  Representations (ICLR)}, 2017.

\bibitem{bowman2015generating}
S.~R. Bowman, L.~Vilnis, O.~Vinyals, A.~M. Dai, R.~Jozefowicz, and S.~Bengio.
\newblock Generating sentences from a continuous space.
\newblock {\em arXiv preprint arXiv:1511.06349}, 2015.

\bibitem{bunel2017learning}
R.~Bunel, A.~Desmaison, M.~P. Kumar, P.~H.~S. Torr, and P.~Kohli.
\newblock Learning to superoptimize programs, 2017.

\bibitem{chechik2010large}
G.~Chechik, V.~Sharma, U.~Shalit, and S.~Bengio.
\newblock Large scale online learning of image similarity through ranking.
\newblock {\em Journal of Machine Learning Research}, 11(3), 2010.

\bibitem{chen2018treetotree}
X.~Chen, C.~Liu, and D.~Song.
\newblock Tree-to-tree neural networks for program translation, 2018.

\bibitem{chen2018tree}
X.~Chen, C.~Liu, and D.~Song.
\newblock Tree-to-tree neural networks for program translation.
\newblock {\em arXiv preprint arXiv:1802.03691}, 2018.

\bibitem{fabius2015variational}
O.~Fabius and J.~R. van Amersfoort.
\newblock Variational recurrent auto-encoders, 2015.

\bibitem{fu2019cyclical}
H.~Fu, C.~Li, X.~Liu, J.~Gao, A.~Celikyilmaz, and L.~Carin.
\newblock Cyclical annealing schedule: A simple approach to mitigating kl
  vanishing.
\newblock {\em arXiv preprint arXiv:1903.10145}, 2019.

\bibitem{grave2017efficient}
E.~Grave, A.~Joulin, M.~Cissé, D.~Grangier, and H.~Jégou.
\newblock Efficient softmax approximation for gpus, 2017.

\bibitem{graves2013hybrid}
A.~Graves, N.~Jaitly, and A.-r. Mohamed.
\newblock Hybrid speech recognition with deep bidirectional lstm.
\newblock In {\em 2013 IEEE workshop on automatic speech recognition and
  understanding}, pages 273--278. IEEE, 2013.

\bibitem{gulwani2017program}
S.~Gulwani, O.~Polozov, R.~Singh, et~al.
\newblock Program synthesis.
\newblock {\em Foundations and Trends{\textregistered} in Programming
  Languages}, 4(1-2):1--119, 2017.

\bibitem{hajipour2019samplefix}
H.~Hajipour, A.~Bhattacharya, and M.~Fritz.
\newblock Samplefix: Learning to correct programs by sampling diverse fixes,
  2019.

\bibitem{holtzman2019curious}
A.~Holtzman, J.~Buys, L.~Du, M.~Forbes, and Y.~Choi.
\newblock The curious case of neural text degeneration.
\newblock {\em arXiv preprint arXiv:1904.09751}, 2019.

\bibitem{kao2020comparison}
C.-C. Kao, M.~Sun, W.~Wang, and C.~Wang.
\newblock A comparison of pooling methods on lstm models for rare acoustic
  event classification, 2020.

\bibitem{kingma2014adam}
D.~P. Kingma and J.~Ba.
\newblock Adam: A method for stochastic optimization.
\newblock {\em arXiv preprint arXiv:1412.6980}, 2014.

\bibitem{kingma2013auto}
D.~P. Kingma and M.~Welling.
\newblock Auto-encoding variational bayes.
\newblock {\em arXiv preprint arXiv:1312.6114}, 2013.

\bibitem{kingma2019introduction}
D.~P. Kingma and M.~Welling.
\newblock An introduction to variational autoencoders.
\newblock {\em arXiv preprint arXiv:1906.02691}, 2019.

\bibitem{koza1992genetic}
J.~R. Koza.
\newblock {\em Genetic programming: on the programming of computers by means of
  natural selection}, volume~1.
\newblock MIT press, 1992.

\bibitem{kullback1951information}
S.~Kullback and R.~A. Leibler.
\newblock On information and sufficiency.
\newblock {\em The annals of mathematical statistics}, 22(1):79--86, 1951.

\bibitem{kusner2017grammar}
M.~J. Kusner, B.~Paige, and J.~M. Hern{\'a}ndez-Lobato.
\newblock Grammar variational autoencoder.
\newblock In {\em International Conference on Machine Learning}, pages
  1945--1954. PMLR, 2017.

\bibitem{le2018maximal}
T.~Le, T.~Nguyen, T.~Le, D.~Phung, P.~Montague, O.~D. Vel, and L.~Qu.
\newblock Maximal divergence sequential autoencoder for binary software
  vulnerability detection.
\newblock In {\em International Conference on Learning Representations}, 2019.

\bibitem{li2015hierarchical}
J.~Li, M.-T. Luong, and D.~Jurafsky.
\newblock A hierarchical neural autoencoder for paragraphs and documents.
\newblock {\em arXiv preprint arXiv:1506.01057}, 2015.

\bibitem{li2018code}
J.~Li, Y.~Wang, M.~R. Lyu, and I.~King.
\newblock Code completion with neural attention and pointer networks, 2018.

\bibitem{liskowski2020program}
P.~Liskowski, K.~Krawiec, N.~E. Toklu, and J.~Swan.
\newblock Program synthesis as latent continuous optimization: evolutionary
  search in neural embeddings.
\newblock In {\em Proceedings of the 2020 Genetic and Evolutionary Computation
  Conference}, pages 359--367, 2020.

\bibitem{long2019preventing}
T.~Long, Y.~Cao, and J.~C.~K. Cheung.
\newblock Preventing posterior collapse in sequence vaes with pooling.
\newblock {\em arXiv preprint arXiv:1911.03976}, 2019.

\bibitem{Luan_2019}
S.~Luan, D.~Yang, C.~Barnaby, K.~Sen, and S.~Chandra.
\newblock Aroma: code recommendation via structural code search.
\newblock {\em Proceedings of the ACM on Programming Languages},
  3(OOPSLA):1–28, Oct 2019.

\bibitem{genimp2}
M.~Monperrus.
\newblock Automatic software repair: a bibliography.
\newblock {\em CoRR}, abs/1807.00515, 2018.

\bibitem{ast}
I.~Neamtiu, J.~S. Foster, and M.~Hicks.
\newblock Understanding source code evolution using abstract syntax tree
  matching.
\newblock In {\em Proceedings of the 2005 international workshop on Mining
  software repositories}, pages 1--5, 2005.

\bibitem{papineni2002bleu}
K.~Papineni, S.~Roukos, T.~Ward, and W.-J. Zhu.
\newblock Bleu: a method for automatic evaluation of machine translation.
\newblock In {\em Proceedings of the 40th annual meeting of the Association for
  Computational Linguistics}, pages 311--318, 2002.

\bibitem{pennington2014glove}
J.~Pennington, R.~Socher, and C.~D. Manning.
\newblock Glove: Global vectors for word representation.
\newblock In {\em Proceedings of the 2014 conference on empirical methods in
  natural language processing (EMNLP)}, pages 1532--1543, 2014.

\bibitem{genimp1}
J.~Petke, S.~O. Haraldsson, M.~Harman, W.~B. Langdon, D.~R. White, and J.~R.
  Woodward.
\newblock Genetic improvement of software: A comprehensive survey.
\newblock {\em IEEE Transactions on Evolutionary Computation}, 22(3):415--432,
  2018.

\bibitem{shin2019program}
R.~Shin, M.~Allamanis, M.~Brockschmidt, and O.~Polozov.
\newblock Program synthesis and semantic parsing with learned code idioms,
  2019.

\bibitem{tai2015improved}
K.~S. Tai, R.~Socher, and C.~D. Manning.
\newblock Improved semantic representations from tree-structured long
  short-term memory networks.
\newblock {\em arXiv preprint arXiv:1503.00075}, 2015.

\bibitem{takahashi2019variational}
H.~Takahashi, T.~Iwata, Y.~Yamanaka, M.~Yamada, and S.~Yagi.
\newblock Variational autoencoder with implicit optimal priors.
\newblock In {\em Proceedings of the AAAI Conference on Artificial
  Intelligence}, volume~33, pages 5066--5073, 2019.

\bibitem{tiobe2017tiobe}
I.~TIOBE.
\newblock Tiobe index.
\newblock {\em Retrieved from Tiobe Index: https://www. tiobe.
  com/tiobe-index}, 2017.

\bibitem{tufano2019learning}
M.~Tufano, J.~Pantiuchina, C.~Watson, G.~Bavota, and D.~Poshyvanyk.
\newblock On learning meaningful code changes via neural machine translation.
\newblock In {\em 2019 IEEE/ACM 41st International Conference on Software
  Engineering (ICSE)}, pages 25--36. IEEE, 2019.

\bibitem{turing1950computing}
A.~M. Turing and J.~Haugeland.
\newblock {\em Computing machinery and intelligence}.
\newblock MIT Press Cambridge, MA, 1950.

\bibitem{winata2018attention}
G.~I. Winata, O.~P. Kampman, and P.~Fung.
\newblock Attention-based lstm for psychological stress detection from spoken
  language using distant supervision.
\newblock In {\em 2018 IEEE International Conference on Acoustics, Speech and
  Signal Processing (ICASSP)}, pages 6204--6208. IEEE, 2018.

\end{thebibliography}
